%% file: 0-main.tex
\pdfoutput=1
\documentclass[letterpaper, 10 pt, conference]{ieeeconf}

\IEEEoverridecommandlockouts                              

\usepackage[backend=biber,
            hyperref=true,
            url=false,
            isbn=false,
            doi=false,
            backref=false,
            style=ieee,
            citestyle=numeric-comp,
            sorting=none, 
            block=none]{biblatex}
        
\renewcommand{\bibfont}{\small}
\input{0-preamble}

\title{\LARGE \bf
\model: Data-Driven Operational Space Control\\for Adaptive and Robust Robot Manipulation
}

\author{Josiah Wong$^{1,2}$, Viktor Makoviychuk$^{2}$, Anima Anandkumar$^{2,3}$, Yuke Zhu$^{2,4}$
\thanks{$^{1}$Stanford University $^{2}$NVIDIA $^{3}$California Institute of Technology $^{4}$The University of Texas at Austin. Correspondance to {\tt\small jdwong@stanford.edu}}
}

\begin{document}
\maketitle


\begin{abstract}
Learning performant robot manipulation policies can be challenging due to high-dimensional continuous actions and complex physics-based dynamics. This can be alleviated through intelligent choice of action space. Operational Space Control (OSC) has been used as an effective task-space controller for manipulation. Nonetheless, its strength depends on the underlying modeling fidelity, and is prone to failure when there are modeling errors. In this work, we propose \underline{OSC} for \underline{A}daptation and \underline{R}obustness (\model), a data-driven variant of OSC that compensates for modeling errors by inferring relevant dynamics parameters from online trajectories. \model decomposes dynamics learning into task-agnostic and task-specific phases, decoupling the dynamics dependencies of the robot and the extrinsics due to its environment. This structure enables robust zero-shot performance under out-of-distribution and rapid adaptation to significant domain shifts through additional finetuning. We evaluate our method on a variety of simulated manipulation problems, and find substantial improvements over an array of controller baselines. For more results and information, please visit \website.

\end{abstract}


\input{1-intro}
\input{3-background}
\input{4-deeposc}
\input{5-experiment}
\input{2-related}
\input{7-conclusion}





\clearpage

\section*{ACKNOWLEDGMENT}
We would like to thank Jim Fan and the NVIDIA AI-ALGO team for their insightful feedback and discussion, and the NVIDIA IsaacGym simulation team for providing technical support.


\renewcommand*{\bibfont}{\footnotesize}
\printbibliography 



\end{document}

%% file: 0-preamble.tex
\usepackage{marginnote}
\usepackage{graphics}
\usepackage[pdftex]{graphicx}
\usepackage{wrapfig}
\DeclareGraphicsExtensions{.pdf,.png,.jpg}

\usepackage[font={small}]{caption}
\usepackage{subcaption}
\usepackage[rightcaption]{sidecap}
\usepackage{pbox}

\usepackage{mathtools}
\usepackage{amsmath, amssymb, amscd}
\usepackage{ wasysym } 
\usepackage{amsfonts}
\usepackage{mathptmx} 
\usepackage{gensymb} 
\usepackage{nicefrac}       

\DeclareMathAlphabet{\mathcal}{OMS}{lmsy}{m}{n}
\DeclareSymbolFont{largesymbols}{OMX}{cmex}{m}{n}
\usepackage{textcomp} 
\usepackage{dsfont}

\usepackage{array} 
\usepackage{tabularx}
\usepackage{multirow}
\usepackage{multicol}
\usepackage{booktabs}

\usepackage[
  separate-uncertainty = true,
  multi-part-units = repeat
]{siunitx}
\usepackage[T1]{fontenc} 
\usepackage[utf8]{inputenc}
\usepackage[english]{babel} 
\usepackage{units}
\usepackage{bm}
\usepackage{times} 
\usepackage{xspace}
\usepackage{flushend}
\usepackage{csquotes}
\usepackage{makeidx}

\let\labelindent\relax
\usepackage{enumitem}

\usepackage{soul} 
\usepackage{subfiles} 

\usepackage[protrusion=true,expansion=true]{microtype}
\usepackage[yyyymmdd]{datetime}

\date{\protect\formatdate{1}{1}{2001}}

\usepackage{url}
\makeatletter
\g@addto@macro{\UrlBreaks}{\UrlOrds}
\makeatother
\usepackage{color}
\usepackage[usenames,dvipsnames,table]{xcolor}
\usepackage[pdfborder={0 0 0.5}]{hyperref}
\hypersetup{
    colorlinks=true,
    linkcolor=black,
    citecolor=black,
    filecolor=cyan,
    urlcolor=black
}

\usepackage{soul} 

\newcommand{\tocite}[1]{%
\textcolor{red}{[cite:\ifthenelse{\equal{#1}{}}{}{#1}?]}
}

\newcommand{\ignore}[1]{}

\newcommand{\model}{\textsc{OSCAR}\xspace}
\newcommand{\delan}{\textsc{DeLaN}\xspace}

\usepackage{flushend}
\usepackage{balance}

\usepackage{algorithm}
\usepackage{algpseudocode}

\renewcommand{\Vec}[1]{\boldsymbol{#1}}
\newcommand{\Hgt}{\Vec{H}}
\newcommand{\Hbase}{\Hat{\Hgt}}
\newcommand{\Hres}{\Tilde{\Hgt}}
\newcommand{\q}{\Vec{q}}
\newcommand{\tor}{\Vec{\tau}}
\newcommand{\qd}{\dot{\q}}
\newcommand{\qdd}{\ddot{\q}}
\newcommand{\ddt}[1][]{\frac{d}{dt}{#1}}

\newcommand{\pddq}[1][]{\frac{\partial}{\partial \q}{#1}}

\newcommand{\g}{\Vec{g}}
\newcommand{\V}{\Vec{V}}

\newcommand{\Lgt}{\Vec{L}}

\newcommand{\Ldiag}{\Lgt_d}

\newcommand{\Lres}{\Tilde{\Lgt}}
\newcommand{\Lresdiag}{\Tilde{\Lgt}_d}
\newcommand{\f}[1][]{#1{f}(\q, \qd, \tor, #1{\Hgt}, #1{\g})}
\newcommand{\finv}[1][]{#1{f}^{-1}(\q, \qd, \qdd, #1{\Hgt}, #1{\g})}
\newcommand{\T}[1]{{#1}^{\top}}
\newcommand{\torque}{\Vec{\tau}}

\newcommand{\latent}{\Vec{z}}
\newcommand{\resmag}{\epsilon}

\newcommand{\R}{\mathds{R}}
\newcommand{\website}{\texttt{https://cremebrule.github.io/oscar-web/}}

%% file: 1-intro.tex
\section{Introduction}
\label{sec:intro}

Robust robot manipulation for real-world tasks is challenging as it requires controlling robots with many degrees of freedom to perform contact-rich interactions that can quickly adapt to varying conditions. While general reinforcement learning algorithms~\cite{haarnoja2018sac, barth2018d4pg, mahmood2018benchmark} can be employed for designing robot controllers from experiences, a crucial and often neglected design choice is the action space for specifying the desired motions of robots~\cite{martin-martin2019vices}.
 Recent work has shown promise in using abstract action representations, rather than low-level torque actuation, for expediting manipulation learning. A broad range of action representations have been examined, including task-space commands~\cite{kalakrishnan2011forcecontrol,lee2019multimodal,martin-martin2019vices,sadeghian2014tasknullspace,xian2004quatfeedback,shuzhi1997adaptivenn}, latent-space action embeddings~\cite{allshire2021laser,li2020usermappings,karamcheti2021latentteleop,pahic2021robotskilllearning}, and high-level goal specifications~\cite{lippi2020visualactionplanning,wang2019visualplanningacting}. These action abstractions have been shown to ease the exploration burdens of reinforcement learning and improve its sample efficiency.
 
 Among these action representations, Operational Space Control (OSC)~\cite{khatib1987osc} has emerged as an effective task-space controller for contact-rich manipulation tasks~\cite{lee2019multimodal,martin-martin2019vices}. OSC parameterizes motor commands by end-effector displacement and maps these commands into deployable joint torques.
It has many advantages with its dynamically consistent formalism, including modeling compliance, compressing the higher-dimensional non-linear $N$-DOF torque control into orthogonal 6-DOF actions, and accelerating learning by enabling agents to reason directly in the task space.

However, OSC's benefits are not always realized. As a model-based controller, its practical strength heavily relies on a high-fidelity model of the \emph{mass matrix}. This time-varying quantity accounts for the robot's mass distribution at its current configuration and is crucial for calculating torques. In the presence of inaccuracies in dynamics modeling, OSC's performance quickly deteriorates~\cite{nakanishi2008oscevaluation}. We illustrate this problem in a simple path tracing task in Fig.~\ref{fig:pull}, where unmodeled extrinsics parameters such as friction and external forces significantly reduce the fidelity of the analytical mass matrix and deteriorate the tracking accuracy. The problem with modeling errors is exacerbated by the fact that controller design is often decoupled from policy learning, and becomes especially pronounced during task transfer settings such as simulation-to-real where there can be significant domain shifts. While a policy can finetune itself, mass matrix quality cannot be improved with data due to its analytical formulation.

\input{fig-pull}

\input{fig-model}

How can we overcome the key limitation of OSC? We observe that its key element, the mass matrix, is subject to physics-based constraints as expressed by robot dynamics differential equations~\cite{niku2001introduction} in classical mechanics. 
Recent advances in physics-informed machine learning has developed a new family of neural networks~\cite{chen2018neuralode, shao2019pinn, li2020fno} that learn such differential equations from data.
In particular, Lutter et al.~\cite{lutter2019delan, lutter2019delan4ec} introduced Deep Lagrangian Networks (\delan) that directly infers the mass matrix from sampled trajectories. However, despite promising results on real robots, \delan is limited by its modeling capacity and underlying assumptions, viz., that robots move in free space with constant mass and no external disturbances. This is unrealistic for most robot manipulation tasks, which require contact interaction with the environment. Hence, we need a more versatile and robust model for contact-rich manipulation tasks.

\textbf{Our Approach:} To this end, we introduce \underline{OSC} for \underline{A}daptation and \underline{R}obustness (\model), a data-driven variant of OSC that leverages a neural network-based  physics model to infer relevant modeling parameters and enable online adaptation to changing dynamics. \model addresses \delan's limitations by extending its formulation to be amenable to general dynamic settings such as robot manipulation.

Concretely, we introduce latent extrinsics inputs that capture task-specific environment factors and robot parameters, and design an encoder to infer these extrinsics from the state-action history. These extrinsics allow our model to infer wide variations in dynamics during training, and robustly work in out-of-distribution settings. Furthermore, we augment the dynamics model with a residual component. This design factorizes the dynamics model into a canonical task-agnostic component learned online from scratch using free-space motion and a constrained task-specific residual, allowing for fast adaptation to dynamics change through residual learning.

We evaluate \model in three diverse manipulation tasks: Path Tracing, Cup Pouring, and Puck Pushing, all of which become especially challenging due to varying degrees of dynamics variations. We evaluate extensive baselines, and find that \model is much more \textbf{stable}, being the only model to achieve task success on all tasks when evaluated on training distributions, \textbf{robust}, exhibiting significantly less policy degradation when evaluated in zero-shot under out-of-distribution, and \textbf{adaptive}, being the only model to reconverge to similar levels of task performance across all tasks when quickly finetuned under significant domain shifts.

%% file: fig-pull.tex
\begin{figure}
\setlength{\fboxrule}{1pt}
\setlength{\fboxsep}{0pt}
\centering
\begin{subfigure}{0.48\textwidth}
\centering
\includegraphics[width=\columnwidth]{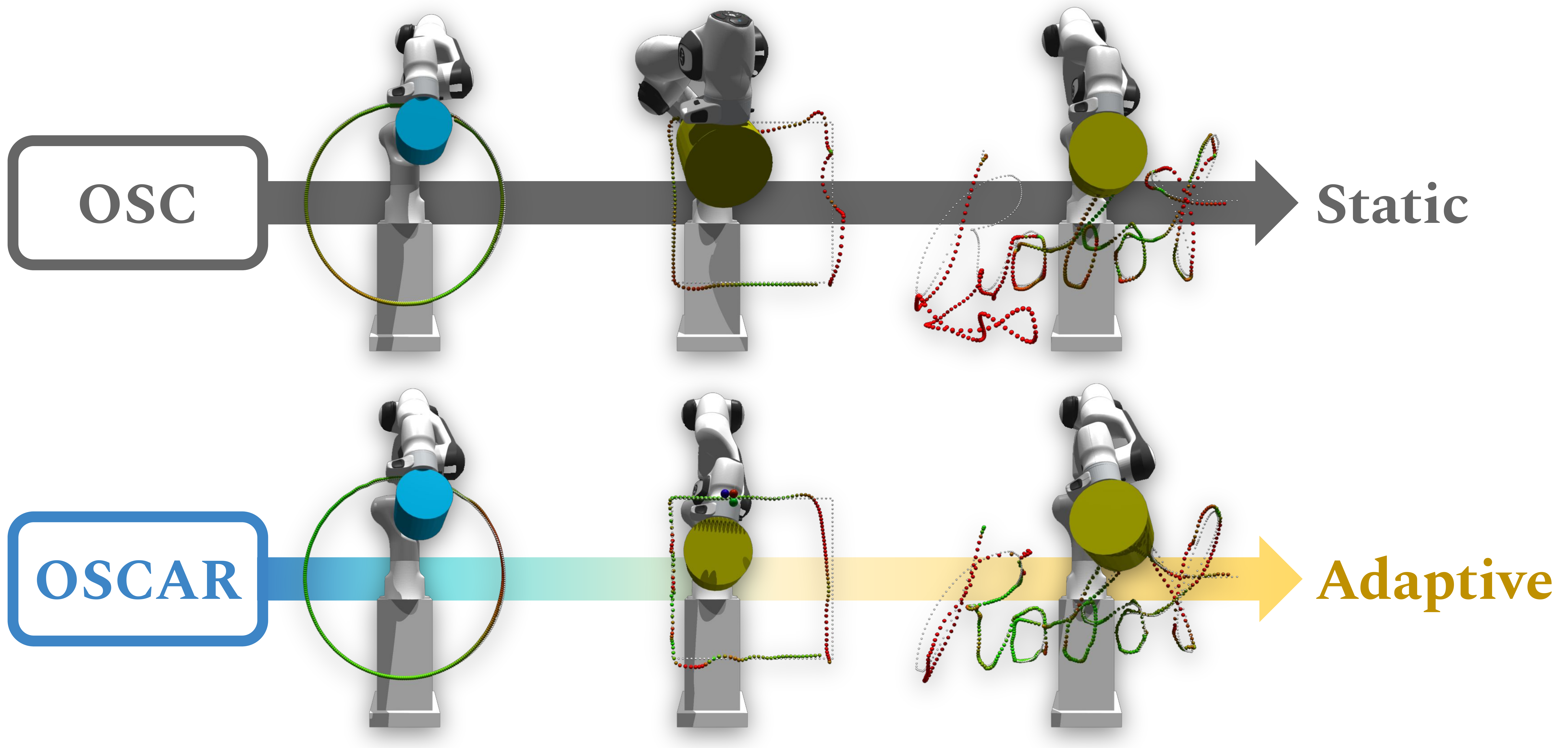}
\end{subfigure}
\caption{\textbf{Adapting to Changing Dynamics.} Path Tracing is an example of a task requiring accurate trajectory motion and can be sensitive to dynamics parameters. For example, while in-distribution performance may be similar across multiple models (left), changing the train distribution end-effector weight (\textit{blue}) and trajectories (\textit{circles}) to unseen values (\textit{yellow}, \textit{squares}) can degrade performance if modeling errors are not accounted for. In contrast, \model directly learns a dynamics representation online from scratch, enabling it to more robustly perform under zero-shot transfer conditions (middle) and quickly finetune under more extreme domain shifts, such as handwritten cursive (right).}
\label{fig:pull}
\vspace{-6pt}
\end{figure}

%% file: fig-model.tex
\begin{figure*}
\centering
\begin{subfigure}{0.96\textwidth}
\centering
\includegraphics[width=\columnwidth]{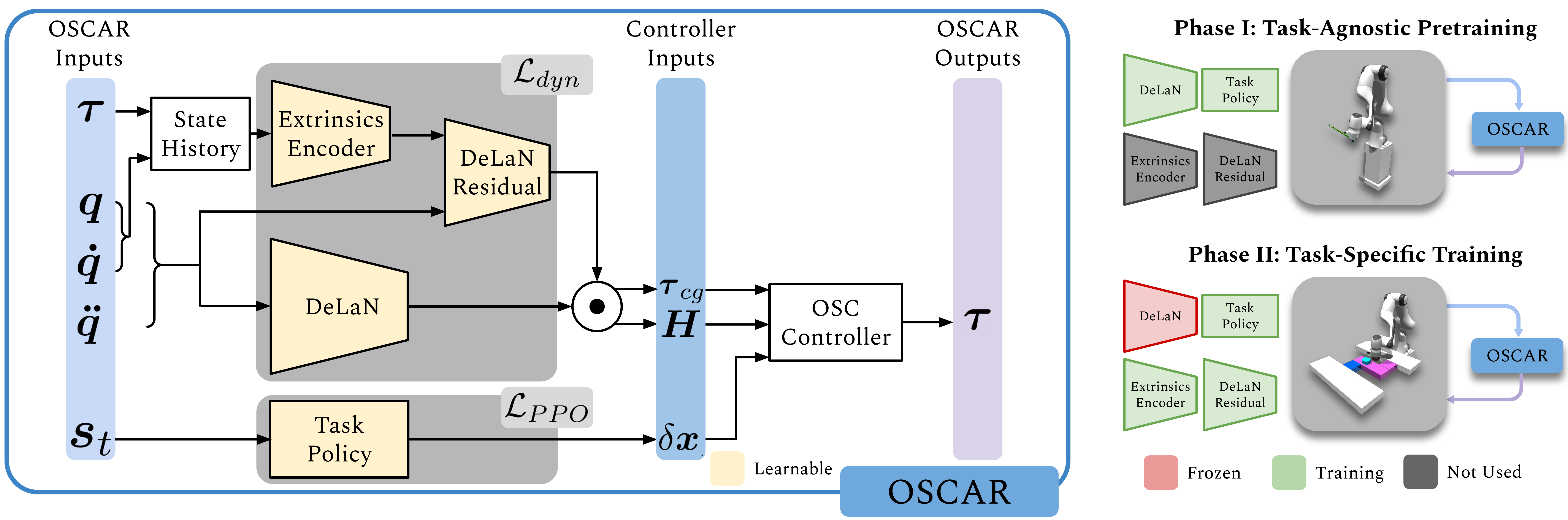}
\end{subfigure}
\caption{\textbf{\model Architecture.} Training \model is split sequentially into a task-agnostic (top right) and task-specific (bottom right) phases. During the task-agnostic phase, an initial dynamics model is bootstrapped using \delan using trajectories outputted by a policy being simultaneously trained to follow random waypoint trajectories. During the task-specific phase, we discard the original policy and freeze the base \delan model, and then finetune the dynamics model using a residual while simultaneously learning a new task-relevant policy. The dynamics model's objective is dyanmics loss $\mathcal{L}_{dyn}$ whereas the policy is trained via RL using PPO loss $\mathcal{L}_{PPO}$.}
\label{fig:model}
\vspace{-6pt}
\end{figure*}

%% file: 3-background.tex
\section{Background}
\label{sec:background}

\subsection{Preliminaries}
We model robot manipulation tasks as an infinite-horizon discrete-time Markov Decision Process (MDP) $\mathcal{M} = \langle \mathcal{S}, \mathcal{A}, \mathcal{T}, \mathcal{R}, \gamma, \rho_0 \rangle$, where $\mathcal{S}$ is the state space, $\mathcal{A}$ is the action space, $\mathcal{T}(s_{t+1} | s_t, a_t)$ is the state transition probability distribution, $\mathcal{R}(s_t, a_t, s_{t+1})$ is the reward function, $\gamma \in [0, 1)$ is the reward discount factor, and $\rho_0(\cdot)$ is the initial state distribution. Per timestep $t$, an agent observes $s_t$, deploys policy $\pi$ to choose an action $a_t \sim \pi(a_t | s_t)$, and executes the action in the environment, observing the next state $s_{t+1} \sim \mathcal{T}(\cdot)$ and receiving reward $r_t = \mathcal{R}(\cdot)$. We seek to learn policy $\pi$ that maximizes the discounted expected return $\mathds{E} [\sum_{t=0}^{\infty} \gamma^t \mathcal{R}(s_t, a_t, s_{t+1})]$.

\subsection{Operational Space Control (OSC)}
OSC is a dynamically consistent controller that models compliant task-space motion. It is particularly useful for robot manipulation, where the robot's end-effector is often the critical task point and must produce compliant behavior for either safety or task-specific reasons. The control law is:

\begin{equation}
    \tor = \Vec{J^\top}_{ee}(\q)[\Vec{\Lambda}_{ee}(\q)[\Vec{k}_p (\Vec{x}_d - \Vec{x}) - \Vec{k}_v ((\Vec{\Dot{x}}_d - \Vec{\Dot{x}}))]] \label{eq:osc}
\end{equation}
where the inertial matrix $\Vec{\Lambda}_{ee} \in \R^{6 \times 6}$ and the Jacobian $\Vec{J}_{ee}$ both specified in the end-effector frame maps the desired PD $\Vec{k}_p, \Vec{k}_v \in \R^6$ control of desired end-effector pose and velocity $\Vec{x}_d, \Vec{\Dot{x}}_d \in \R^6$ to joint-space control torques $\tor \in \R^N$, where $N$ is the number of robot joints. $\Vec{k}_p$ and $\Vec{k}_v$ specify the relative compliance of the controller~\cite{khatib1987osc}. These values can be either defined \textit{a priori} or learned as part of the action space (VICES~\cite{martin-martin2019vices}). For VICES, the action space is $\R^{18}$, instead of the usual $\R^6$ with $\Vec{\Dot{x}}_d = 0$. This choice of action space has been shown to be especially advantageous for contact-rich manipulation~\cite{luo2019rl_vices}. For this reason, we apply VICES to all OSC-based models unless otherwise noted.

Crucially, $\Vec{\Lambda}_{ee}$ depends on the \textit{mass matrix} $\Hgt(\q)$, which in turn depends on the joint state $\q$ and potentially time if the overall system mass is time-varying. Inaccurate modeling of the mass matrix can result in unstable controller behavior. This problem is precisely the challenge that we seek to address through learning $\Hgt$ online from sampled trajectories.

\subsection{Deep Lagrangian Networks (\delan)}
Lutter et al.~\cite{lutter2019delan, lutter2019delan4ec} have shown that arbitrary rigid body dynamics can be inferred from free space motion via the Lagrangian mechanics, captured by the following forward and inverse dynamics equations:
\begin{equation}
    \f = \Hgt^{-1} \left( \tor -\Dot{\Hgt}\qd + \frac{1}{2} \T{\left(\pddq((\T{\qd} \Hgt \qd)\right)} -\g \right) \label{eq:lagrangian_for}
\end{equation}
\begin{equation}
    \finv = \Hgt \qdd + \ddt (\Hgt) \qd - \frac{1}{2} \T{(\pddq(\T{\qd} \Hgt \qd))} + \g \label{eq:lagrangian_inv}
\end{equation}
where $\q$, $\qd$, and $\qdd$ are the system's generalized coordinates and their first- and second-order time derivatives, respectively, and $\g$ is the time-varying sum of potential forces. \delan parameterizes unknown functions $\Hgt, \g$ with neural networks.

Given known values $\q_t$, $\qd_t$, $\qdd_t$, \delan predicts current torques $\tor_{pred}$ using Eq. \ref{eq:lagrangian_inv}, and additionally uses observed torques $\tor$ to predict joint accelerations $\qdd_{pred}$ using Eq. \ref{eq:lagrangian_for}. \delan is trained with L2 regression loss with respect to known observed quantities, and includes forward, inverse, and energy conservation dynamics losses (Fig \ref{fig:delan}).

In order to allow efficient off-the-shelf back-propagation, the loss computations require analytical first-order derivative computations of $\pddq[\Hgt]$. Since $\Hgt$ is parameterized by a neural network with $\q$ as inputs, this can be achieved by exclusively using fully-connected layers (``Lagrangian Layers'') which allow analytical first-order partial derivatives to be computed during the forward pass. 

%% file: 4-deeposc.tex
\section{Our Method (\model)}
\label{sec:deeposc}

\input{fig-delan}

We now propose \model, a data-driven variant of OSC capable of leveraging the strengths of its controller formulation while removing the modeling burden of the high-fidelity and pre-defined mass matrix. We achieve this by learning a dynamics model that directly infers the mass matrix from online trajectories. Crucially, our dynamics model builds upon \delan and concretely improves its modeling capacity by (a) introducing latent extrinsics inputs that capture task-specific environment factors and robot parameters, which can be inferred directly from the state-action history, and (b) augmenting the dynamics model with a residual component, enabling fast adaptation to dynamics changes through residual learning (Fig. \ref{fig:model}). Together, \model enables end-to-end joint learning of the dynamics model and the policy, and can adapt to out-of-distribution environment variations. We describe each of these key features below.

\subsection{Capturing Latent Extrinsics}
\label{sec:latent}
A key limitation of \delan resides in its modeling capacity, which is bottlenecked by its dependency exclusively on $\q$, $\qd$, and $\qdd$. While these values may be sufficient to model dynamics for a robot moving through free space in steady state, it can be problematic for settings where these values are not constant, such as during extended external impedance like grasping or applying the same model across a fleet of robots with varying individual dynamical properties.

\model addresses this problem by implicitly inferring relevant extrinsic parameters directly from a robot's state-action history via learned latent embeddings. This follows the intuition that states are jointly dependent on both actions and underlying dynamics. Thus, learning correlations between recent states and actions can be a viable proxy for inferring extrinsic variations. This hypothesis has been validated by recent work \cite{kumar2021rma} that has shown a similar method robust enough for direct simulation-to-real transfer.

To this end, we add an additional module to our dynamics model, which takes in the robot's recent joint states $\q$, $\qd$ and low-level torques $\torque$, and generates low-dimensional latent embeddings $\latent$ that are directly fed into the \delan residual (Sec. \ref{sec:residual}). Because the base network is restrained to inferring relevant dynamics from $\q$, augmenting inputs with $\latent$ can increase the fidelity of the learned dynamics by allowing additional information flow from the state-action history. We use a shallow 4-layer MLP network as our encoder.

In order to leverage automatic differentiation for calculating the dynamics loss, we must analytically compute the first-order partial derivative $\pddq[\latent]$. However, we crucially assume that these latent embeddings capture environment extrinsics that are \textit{independent} of $\q$, such that $\pddq[\latent] = \Vec{0}$. This makes forward and gradient computations much simpler and also encourages the learned embeddings to be agnostic to spurious correlations with $\q$.

\subsection{Residual-Augmented Mass Matrices}
\label{sec:residual}
Under domain shifts that substantially change the system dynamics, it would be advantageous for a pretrained but possibly inaccurate dynamics model to quickly adapt in an online fashion. Therefore, \model decouples dynamics learning into task-agnostic and task-specific components, inferring an initial reference estimate of the dynamics that can be applied to many models and finetuned on task-specific dynamics. We achieve this by decomposing the learned mass matrix model $\Hgt$ into a \textit{task-agnostic} base $\Hbase$ and a small \textit{task-specific} residual $\Hres$. Learning the task-agnostic base function mitigates the original modeling burden of OSC, and learning a task-specific residual can reduce the domain shift problem by adapting to the current dynamics.

We formulate our residual component $\Hres$ as a multiplicative residual outputting constrained scaling factors such that $\Hgt$ is the element-wise multiplication of $\Hbase$ and $\Hres$:
\begin{equation}
    \Hgt = \Hbase \odot \Hres(\cdot; \phi), \quad ||\Hres - 1||_{\infty} < \resmag, \quad |\resmag| \: \text{small} \label{lmm}
\end{equation}
where $\Hres$ is parameterized by a multi-layered neural network similar to \delan, $\phi$ are the learned network weights, and $\odot$ is the element-wise multiplication operator. We choose this multiplicative structure instead of an additive one because we observe that the individual mass matrix elements span multiple orders of magnitude, thereby increasing the optimization difficulty for additive residuals whose outputs must similarly span such a large range of values. 

We provide joint states $\q$, base estimate $\Hbase$, and latent extrinsics $\latent$ (Sec. \ref{sec:latent}) as the residual's inputs. $\latent$ is included because we observe that environment extrinsics are often unique to the task at hand, and do not necessarily generalize across domains. For this same reason, we choose to exclude the latent extrinsics dependency during the $\Hbase$ pretraining process, which is meant to capture task-agnostic information about the mass matrix.

\input{fig-tasks}

To enforce the residual's bounded influence, the residual neural network's output is passed sequentially through a scaled Tanh and Exponential layers. In this way, we can achieve a residual that is centered at 1 (no change to the base model) with symmetric log scale bounds.

A final requirement is to ensure that both the positive definiteness of $\Hgt$ is preserved, and that $\Hgt$'s first-order partial derivative $\pddq[\Hgt]$ can be analytically computed so that off-the-shelf auto-differentiation packages can still be used. The former is achieved by using a similar decomposition as $\Hbase$ and setting $\Hres = \Lres + \T{\Lres} + \Lresdiag$, where $\Lres$ is a lower diagonal matrix with zeros along the diagonal and $\Lresdiag$ is a sparse matrix with only non-zero elements along its diagonal. For the latter, $\pddq[\Hgt]$ can be computed from Equation \eqref{lmm} using chain rule:

\begin{equation}
    \pddq[\Hgt] = \pddq[\Hbase] \odot \Hres + \Hbase \odot \left(\pddq[\Lres] + \pddq[\T{\Lres}] + \pddq[\Lresdiag]\right) \label{eq:dhdq}
\end{equation}

By defining $\Lres$ and $\Lresdiag$ as Lagrangian Layers~\cite{lutter2019delan}, we can tractably compute each of their first-order partial derivatives $\pddq[(\cdot)]$ directly during the forward pass. We train our model using the same dynamics loss as \delan (Fig. \ref{fig:delan}).

\subsection{Training \model} We seek to decompose our model into a task-agnostic core for maximizing modeling efficiency and a task-specific residual for improved per-task fidelity. We achieve this by splitting learning into pretrain and train phases:

\textit{Task-Agnostic Pretraining:} We first learn an initial estimate of the mass matrix $\Hbase$ from scratch by training the core \delan model in conjunction with a policy trained follow randomly generated straight line trajectories in free space. As this is intended to be a reference network, we do not utilize environment randomization nor end-effector masses in order to decouple these influences from initial dynamics learning.

\textit{Task-Specific Training:} After pretraining, we discard the original policy and freeze the core \delan backbone. For a new task, we proceed to train our residual extrinsics-aware model in conjunction with a new policy to learn residual mass matrix $\Hres$ to better capture the task-specific dynamics.

Our dynamics model runs at the same policy frequency (20Hz), and uses residual scaling limit of $\resmag = 0.1$.

%% file: fig-delan.tex
\begin{figure}
\setlength{\fboxrule}{1pt}
\setlength{\fboxsep}{0pt}
\centering
\begin{subfigure}{0.48\textwidth}
\centering
\includegraphics[width=\columnwidth]{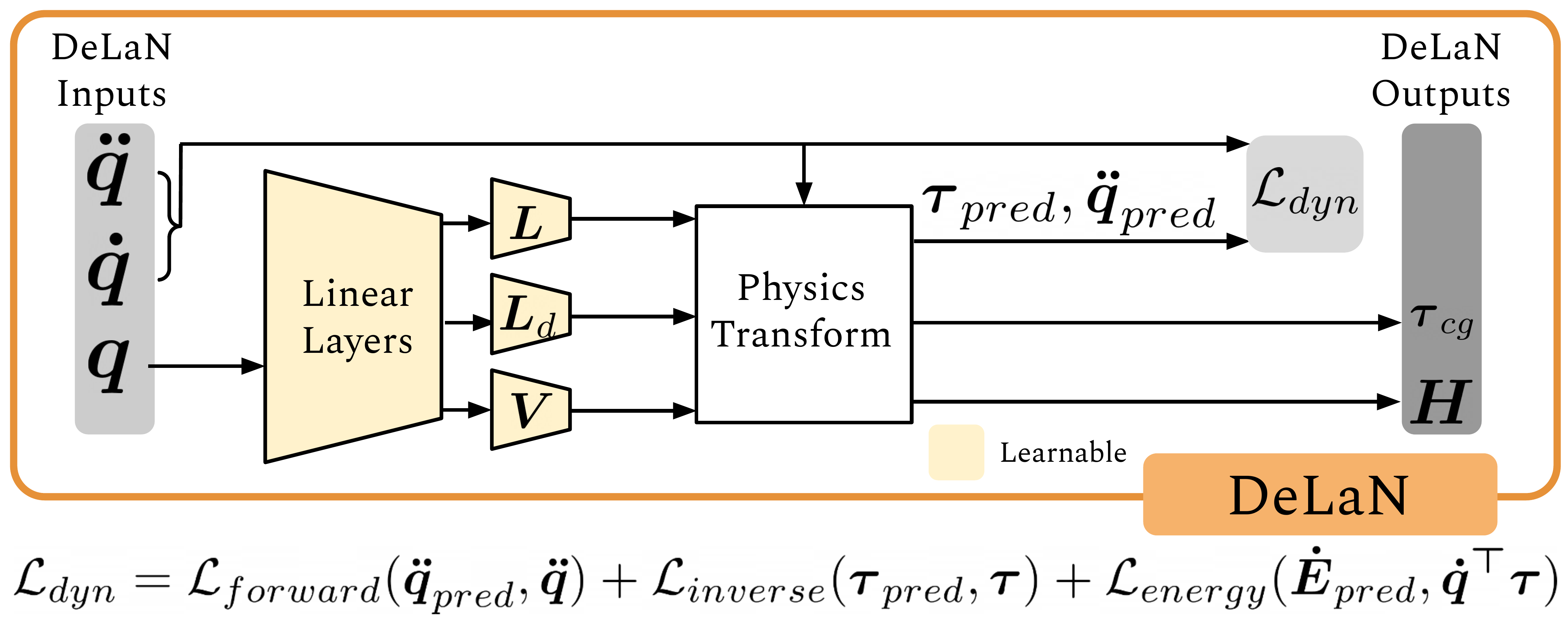}
\end{subfigure}
\caption{\textbf{\delan Architecture.} \delan takes robot joint states $\q$, $\qd$, and $\qdd$ as inputs, and outputs coriolis-gravity compensation torques $\tor_{cg}$ and mass matrix $\Hgt$. A shared linear layer core utilizes separate output heads to bootstrap different dynamics parameters: $\Hgt = \Lgt + \T{\Lgt} + \Ldiag$, and $\g = \pddq[\V]$. These values are manipulated using forward (Eq. \ref{eq:lagrangian_for}) and inverse (Eq. \ref{eq:lagrangian_inv}) dynamics equations to generate predicted values used in the dynamics loss $\mathcal{L}_{dyn}$.}
\label{fig:delan}
\vspace{-12pt}
\end{figure}

%% file: fig-tasks.tex
\begin{figure}
\setlength{\fboxrule}{1pt}
\setlength{\fboxsep}{0pt}
\centering
\begin{subfigure}{0.48\textwidth}
\centering
\includegraphics[width=\columnwidth]{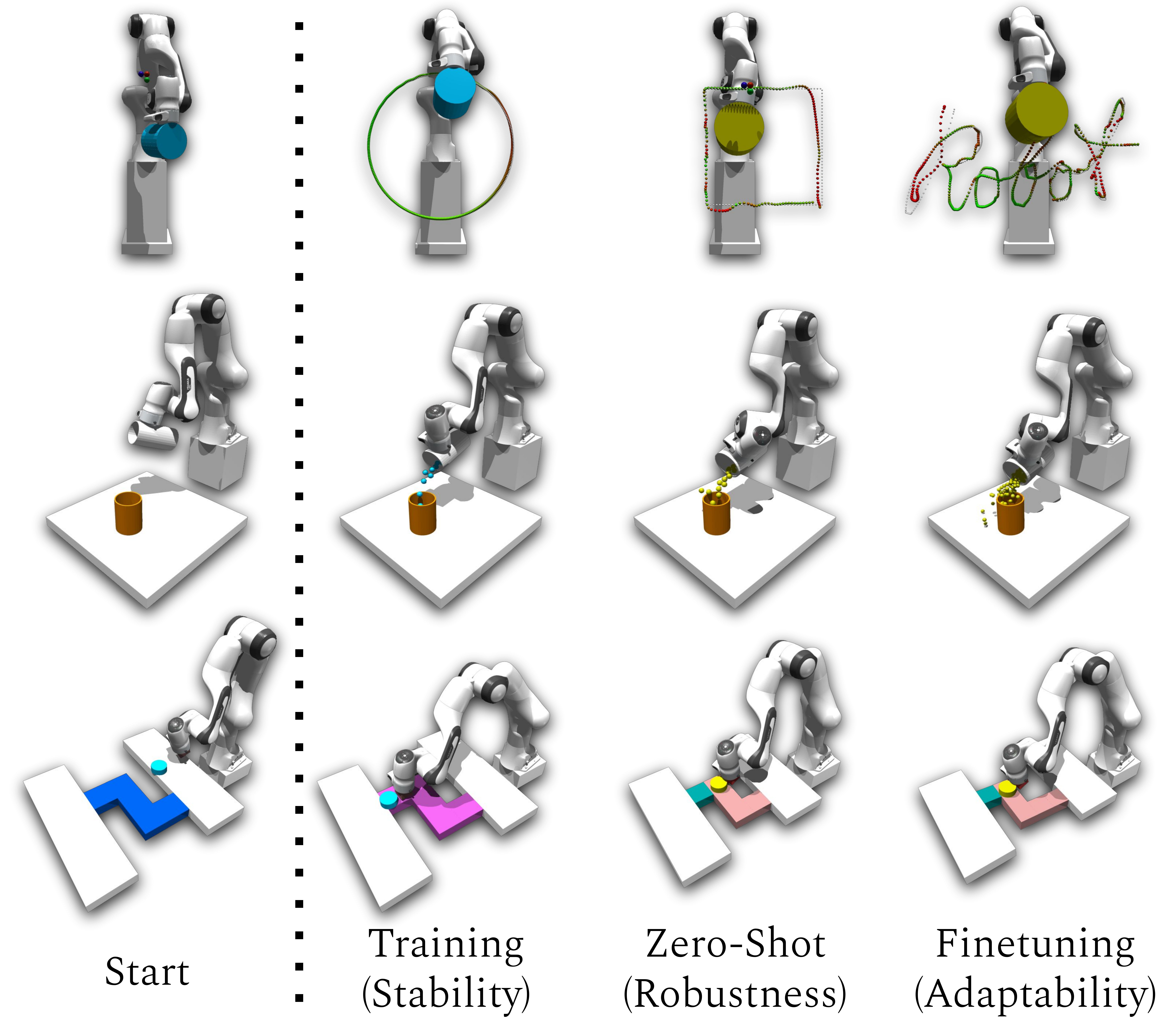}
\end{subfigure}
\caption{\textbf{Tasks.} We present 3 manipulation tasks requiring both dexterity and precision: Path Tracing (top), Cup Pouring (middle), and Puck Pushing (bottom). Dynamics parameters such as mass and friction are randomized, and these tasks require compliance in order to successfully adapt to each instance. We evaluate three variants of each task: the training distribution to evaluate model stability (left), the out-of-distribution configuration with previously unseen dynamics to evaluate zero-shot robustness (middle), and significant out-of-distribution configurations to evaluate adaptability through finetuning (right). We symbolize some relevant parameter changes applied with distinguishing colors and visual changes.}
\label{fig:tasks}
\vspace{-15pt}
\end{figure}

%% file: 5-experiment.tex
\section{Experiments}
\label{sec:exp}

\input{fig-curves}

\subsection{Experimental Setup}

\textbf{Tasks.} We propose a set of simulated manipulation tasks that highlight the benefits of robust compliant control (Fig. \ref{fig:tasks}). All experiments are conducted using the Franka Panda robot. We implement our environments using Isaac Gym~\cite{mackoviychuk2021isaacgym}, a high-fidelity simulator that has been shown to enable simulation-to-real transfer learning to physical hardware~\cite{chebotar2018sim, narang2021sim2real}.

\textit{Path Tracing:} The robot must follow a set of procedurally generated way-points defining a parameterized path. Sampled paths during training include circles of varying sizes with randomized positions and orientations. The robot also has a sealed container attached to its end-effector containing an unknown mass of randomized weight. Task observations include agent end-effector and goal poses. This task focuses on the robot's ability to accurately follow desired trajectories while being robust to varying impedance.

\textit{Cup Pouring:} The robot holds a pitcher containing small particles and must carefully pour them into a cup located on the table while minimizing the amount of spillage. The cup's diameter, height, and position are randomized between episodes. Task observations include end-effector pose, pitcher tilt, cup position, cup radius, cup height, and the proportion of particles both in and outside of the cup. This task focuses on the robot's ability to respond to changing dynamics.

\textit{Puck Pushing:} The robot must carefully push a puck between tables while avoiding knocking the puck off the platform. Task observations include end-effector pose, puck position and tilt, relative puck position to next path waypoint, and the proportion of the path completed. This task focuses on the robot's ability to adapt to sustained external contact.

In all of the tasks, the robot's initial pose, joint friction, damping, armature, and minimum inertia are randomized between episodes. Specific parameters and reward functions can be found on our website.

\input{tab-results_zero_shot}

\textbf{Baselines.} We compare \model against several baseline controllers, including joint-space options Joint Position, Joint Velocity, Joint Torque and task-space options Inverse Kinematics (IK), OSC (no VICES), OSC.  Among these controllers, the joint-space controllers must operate in the highly nonlinear joint actuation, and IK cannot model compliance. 

The torque-based controller baselines are automatically provided with rough estimations of gravitational compensation torques $\tau_{cg} = \sum_{i=1}^{N} \T{\Vec{J}_{i}} m_{i}$, where $\Vec{J}_{i}$ is the Jacobian and $m_{i}$ is the mass for the $i$-th robot link. For \model, $\tau_{cg}$ is generated directly from the learned dynamics model. In contrast, the IK, Joint Velocity, and Joint Position controllers do not operate in torque space, and so we disable gravity for these controllers. They have a distinct advantage over the torque-based controllers because their learned policies do not have to account for the dynamics influences from gravity. OSC baselines are provided the analytical mass matrix. whereas \model does not have access to any analytical dynamics parameters from the robot model.

\subsection{Performance: Train-Distribution Evaluation} We first consider the normal training performance of each controller on each task. We train our policies using PPO~\cite{schulman2017ppo} with 2048 parallelized environments using 3 random seeds, and utilize identical hyperparameters across all models for fair comparison (Fig. \ref{fig:tasks_train}). We find that OSC is often the strongest baseline across all tasks and consistently outperforms OSC without VICES. This result validates OSC's strengths as a task-space compliant controller and highlights the advantages of dynamically choosing compliance gains to enable policy adaptation to changing dynamics.

\model outperforms all baselines across all tasks, with significant improvements in the more complex Cup Pouring (over $\mathbf{90\%}$) and Puck Pushing tasks (over $\mathbf{60\%}$) compared to the next best baseline. This result indicates that policy learning is not hindered by the \model's learned dynamics model, and in fact benefits from the learned task-specific dynamics. We also note that while the decomposition of $\tor_{cg}$ is not supervised during training, the learned values are sufficiently stable for \model to succeed at each task and still significantly outperform other baselines.

\subsection{Robustness: Out-of-Distribution Zero-Shot Evaluation}
After training, we examine each trained model's robustness to novel task instances in a zero-shot setting. We modify each task's parameters to out-of-distribution combinations previously unseen during training. We maximize the influence of these extrinsics by setting dynamics parameters to their maximum training range values, and significantly increasing relevant masses over the training maximum value. We measure relevant success metrics for each task, and find that \model is the most robust to policy degradation, outperforming the next best baseline by $\mathbf{63\%}$ on the Puck Pushing task and $\mathbf{3780\%}$ on the Cup Pouring task (Table \ref{tab:results_zero_shot}). These results show that \model can be readily deployed in previously unseen dynamics while maintaining high task performance.

\subsection{Adaptability: Out-of-Distribution Finetuning}
Finally, we evaluate each model's ability to rapidly adapt to distribution shifts. In this experiment, we aggressively modify the task parameters and then allow each trained model to finetune under the new settings. To reflect more realistic transfer settings with limited resource bandwidth, we train using only 4 parallel environments instead of 2048 as before. We find that \model re-converges much more quickly compared to all other baselines and also achieves over $\mathbf{500\%}$ for Cup Pouring and $\mathbf{90\%}$ for Puck Pushing performance improvements over the next best baseline (Fig. \ref{fig:tasks_train}). These results suggest that \model helps overcome the exploration burden for difficult task instances by leveraging a previously trained model and further adapting it online. This property is appealing for task transfer, where \model can leverage the advantages of large-scale simulations while quickly adjusting to new environment conditions in a sample-efficient manner.

\subsection{Ablation Study}
Lastly, we compare \model against multiple ablative self-baselines on the Cup Pouring task to validate some key design decisions. We report both the training distribution performance and zero-shot out-of-distribution degradation (Table \ref{tab:ablation}). We find that while our model is marginally outperformed by some other variant during training, it is the most robust under zero-shot domain shift and results in $\mathbf{34.7\%}$ less degradation compared to the next strongest ablative model. These results validate the key modeling components of \model: namely, the multiplicative residual (vs. additive), extrinsics encoder (vs. no extrinsics), and two-step task-agnostic task-specific method used (vs. using no residual and manipulating the base \delan structure instead).

\input{tab-ablation}

%% file: fig-curves.tex
\begin{figure}
\centering
\begin{subfigure}{\columnwidth}
\centering
\includegraphics[width=\columnwidth]{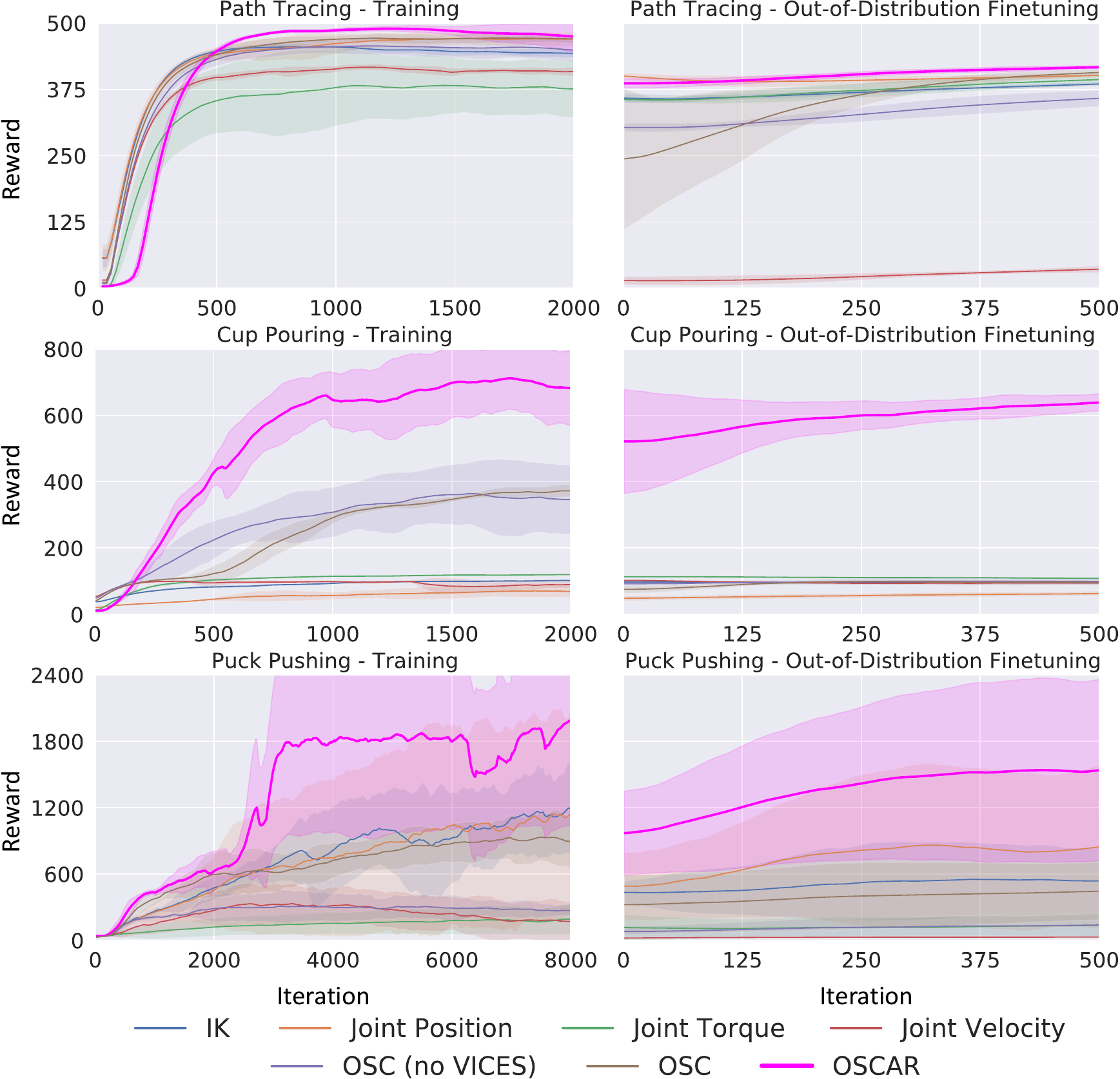}
\end{subfigure}
\caption{\textbf{Task Results.} (Left) When initially trained on each task, \model already outperforms the strongest baselines without leveraging any privileged model information such as mass matrix $\Hgt$ used by analytical OSC baselines during training. (Right) When  finetuned under significant domain shifts, \model re-converges more quickly to more performant levels compared to all the baselines.}
\vspace{-1mm}
\label{fig:tasks_train}
\end{figure}

%% file: tab-results_zero_shot.tex
\begingroup
\setlength{\tabcolsep}{2pt} 
\begin{table}
  \caption{\textbf{Zero-Shot Generalization Results}}
  \label{tab:results_zero_shot}
  \centering
  \begin{tabular}{lccc}
    \toprule
    Model & \begin{tabular}[c]{@{}c@{}}Path Tracing\\MSE (mm)\end{tabular} & \begin{tabular}[c]{@{}c@{}}Cup Pouring\\\% Filled\end{tabular} & \begin{tabular}[c]{@{}c@{}}Puck Pushing\\\% Completed\end{tabular} \\
    \midrule
    Joint Position & $24.4 \pm 34.2$ & $1.9 \pm 2.7$ & $22.8 \pm 16.1$ \\
    Joint Velocity & $482.2 \pm 175.4$ & $0.0 \pm 0.0$ & $0.9 \pm 0.5$ \\
    Joint Torque & $37.5 \pm 10.8$ & $0.0 \pm 0.0$ & $8.4 \pm 5.9$ \\
    IK & $27.0 \pm 6.1$ & $0.0 \pm 0.0$ & $33.4 \pm 1.1$ \\
    OSC (no VICES) & $44.6 \pm 7.4$ & $0.4 \pm 0.4$ & $7.7 \pm 2.1$ \\
    OSC & $26.1 \pm 5.3$ & $0.6 \pm 0.2$ & $46.5 \pm 5.3$ \\
    \textsc{OSCAR} (Ours) & $\mathbf{20.0 \pm 2.3}$ & $\mathbf{73.8 \pm 9.2}$ & $\mathbf{76.0 \pm 14.2}$ \\
    \midrule
    $\Delta$ over the Best Baseline & $\mathbf{+18\%}$ & $\mathbf{+3780\%}$ & $\mathbf{+63\%}$ \\
    \bottomrule
  \end{tabular}
\vspace{-15pt}
\end{table}
\endgroup

%% file: tab-ablation.tex
\begingroup
\setlength{\tabcolsep}{5pt} 
\begin{table}
  \caption{\textbf{Ablation Study on Cup Pouring Task}}
  \centering
  \begin{tabular}{lcc}
    \toprule
    Variant & \begin{tabular}[c]{@{}c@{}}Train\\Return\end{tabular} & \begin{tabular}[c]{@{}c@{}}Zero-shot\\Degradation\end{tabular} \\
    \midrule
    \model (Ours) & $844 \pm 19$ & $\mathbf{-62}$ \\
    Additive residual & $813 \pm 24$ & $-109$ \\
    No extrinsics & $\mathbf{845 \pm 45}$ & $-171$ \\
    No residual + finetune pretrain & $711 \pm 76$ & $-166$ \\
    No residual + freeze pretrain & $774 \pm 99$ & $-105$ \\
    No residual + no pretrain & $760 \pm 20$ & $-95$ \\
    \midrule
    $\Delta$ over the Best Baseline & $-0.1\% $ & $\mathbf{+34.7\%}$ \\
    \bottomrule
  \end{tabular}
\label{tab:ablation}
\vspace{-15pt}
\end{table}
\endgroup

%% file: 2-related.tex
\section{Related Work}
\label{sec:related}

\textbf{Action Space for Learning Robot Control.} Choosing an appropriate action space for learning methods in robotics can be nontrivial. While joint-space commands can be immediately deployed on a robot, they are high-dimensional and nonlinear. Recent work has demonstrated that the abstraction provided by task-space controllers, such as inverse kinematics (IK)~\cite{goldenberg1985ik} and operational space control (OSC)~\cite{khatib1987osc}, can improve policy performance and reduce sample complexity~\cite{lee2019multimodal,martin-martin2019vices,zhu2020robosuite, luo2019rl_vices}.  
Nonetheless, these approaches are heavily model-based and require parameter tuning to overcome kinematic redundancies and modeling errors~\cite{dsouza2001learningik, peters2008learningosc}. Despite OSC's more expressive formulation that models compliance, prior work has primarily focused on improving IK~\cite{pannawit2017nn_ik, karlik2000ik_improve, chen2015ik_screw, koker2013ik_errmin}. A few works exploring OSC have sought to learn its parameters in data-driven ways, such as applying reinforcement learning (RL) to track desired trajectories~\cite{peters2007rlosc} or learning the impedance gains for adapting to task-specific dynamics~\cite{martin-martin2019vices}. Unlike these methods which are limited to gain tuning and require knowledge of the underlying dynamics, we seek to completely eliminate the modeling burdens and infer dynamics from trajectories.

\textbf{Deep Learning for Dynamical Systems.} In contrast to policy learning methods that seek to \textit{adapt} to environment dynamics, a parallel line of work has explored directly modeling these system dynamics as learnable ordinary differential equations (ODEs) with deep neural networks. While the majority of these works seek to model arbitrary physical dynamics~\cite{lutter2019delan, lutter2019delan4ec, finzi2020chnn, zhong2021contacthnn}, some have explored specific complex physical phenomena such as contact~\cite{hochlehnert2021contact, pfrommer2020contactnets, parmar2021contact_challenges} and fluid dynamics~\cite{wessels2020nn_fluid, portwood2019nn_turbulence}. These works leverage strong physical priors to enforce model plausibility. While these works have demonstrated promising results in restrictive domains, they have yet to show success for realistic contact-rich dynamics, such as high degree-of-freedom robot arms physically interacting with objects. Our work builds upon Lutter et al.~\cite{lutter2019delan,lutter2019delan4ec}'s \delan model. We augment its modeling capacity to perform joint policy and dynamics learning end-to-end for contact-rich manipulation tasks.

\textbf{System Identification in Robotics.} Our work is also relevant to methods for system identification (sysID), which seek to estimate accurate models of dynamics from data. Classical methods on sysID~\cite{melsa1971system, astrom1971sysid_survey, ljung1998system} define a priori models and seek to estimate its unknown parameters from sampled data. More recent data-driven methods have relaxed the modeling burdens and proposed to train deep neural networks to capture relevant task dynamics parameters either directly~\cite{chebotar2018sim, yu2019sim} or via latent representations~\cite{cully2014adapt,kumar2021rma}. Recent work on simulation-to-real transfer has focused on estimating distributions of simulated dynamics parameters~\cite{ramos2019bayessim} or allowing end-to-end differentiation through simulation~\cite{freeman2021brax, heiden2021disect} for tuning the simulation model's fidelity. In a similar vein as these works, we leverage simulation with varied dynamics to allow our learned dynamics model to infer both direct and latent representations of these parameters. However, in contrast to most of these works that infer relevant parameters in an end-to-end fashion, we decouple the process into task-agnostic and task-specific phases, allowing our model to generate a reference representation that can be tuned to specific tasks.

%% file: 7-conclusion.tex
\section{Conclusion}
We showcased \model, a data-driven method online learning of dynamics parameters for OSC from scratch, removing OSC's modeling burden and accelerating simultaneous end-to-end training of a task policy and a dynamics model. We showed that compared to extensive baselines, \model shows better performance during training, less degradation in zero-shot out-of-distribution, and better adaptation when finetuned under significant domain shifts. While we were only able to validate \model's sim2sim transfer due to limited hardware access during the pandemic, we hope to soon evaluate on sim2real transfer and further develop our method as a means to deploying data-driven OSC control to real robots.